%% file: main.tex
\documentclass[10pt,twocolumn,letterpaper]{article}

\usepackage{cvpr}              

\input{preamble}

\definecolor{cvprblue}{rgb}{0.21,0.49,0.74}
\usepackage[pagebackref,breaklinks,colorlinks,allcolors=cvprblue]{hyperref}
\usepackage{algorithm,algorithmic}
\usepackage{pifont}
\usepackage{etoc}

\def\paperID{6058} 
\def\confName{CVPR}
\def\confYear{2026}
\def\papername{tttLRM\xspace}

\title{\papername: Test-Time Training for Long Context \\ and Autoregressive 3D Reconstruction}

\author{
Chen Wang$^{1*}$ \quad 
Hao Tan$^{2}$ \quad Wang Yifan$^{2}$ \quad Zhiqin Chen$^{2}$ \\
\quad Yuheng Liu$^{3*}$ \quad Kalyan Sunkavalli$^{2}$ \quad Sai Bi$^{2}$ \quad 
Lingjie Liu$^{1\dagger}$ \quad 
Yiwei Hu$^{2\dagger}$ \\
\vspace{3pt}
$^1$University of Pennsylvania \quad 
$^2$Adobe Research \quad $^3$UCI \\
\vspace{3pt}
\url{https://cwchenwang.github.io/tttLRM} 
}

\begin{document}
\renewcommand{\thefootnote}{\fnsymbol{footnote}}
\maketitle

\begin{strip}
    \centering
    \vspace{-5em}
    \centering
    \includegraphics[width=\linewidth]{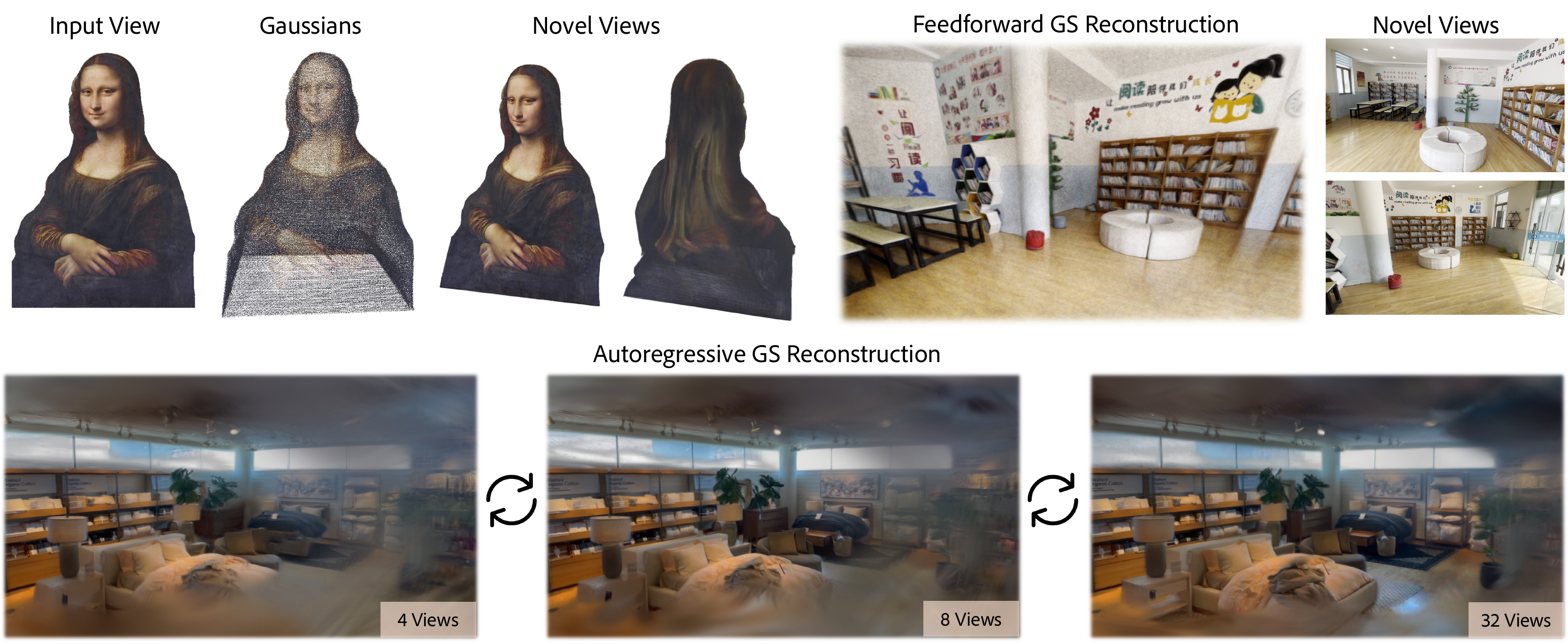}
    \vspace{-15pt}
    \captionof{figure}{We propose \papername, a Large Reconstruction Model based on Test-Time Training, enabling high-resolution, long-context, autoregressive 3D reconstruction. Our model achieves 1) high-resolution (1024px) single-image-to-3D reconstruction via a multi-view generator 2) long-context (64 input views) and feedforward 3DGS reconstruction, and supports 3) autoregressive\protect\footnotemark[3]\ streaming reconstruction.}
    \vspace{-10pt}
    \label{fig:teaser}
\end{strip}

\def\thefootnote{*}\footnotetext{Work done as interns at Adobe Research. $\dagger$Equal advising.}
\def\thefootnote{$\ddagger$}\footnotetext{To clarify, we use autoregressive to denote online, causal version.}

\input{sec/0_abstract}
\input{sec/1-intro}
\input{sec/2-related-work}
\input{sec/3-method}
\input{sec/4-experiments}
\input{sec/5-conclusion}

\section*{Acknowledgment}
The authors would like to thank Ziwen Chen for the evaluation of the baselines and Tianyuan Zhang for helpful discussions on LaCT.

{
    \small
    \bibliographystyle{ieeenat_fullname}
    \bibliography{main}
}
\appendix
\input{sec/X_suppl}

\end{document}

%% file: preamble.tex
\usepackage{cuted}

\def\papername{xxx\xspace}
\newcommand{\topic}[1]
{
\vspace{1pt}\noindent\textbf{#1}
}

\newcommand{\bx}{\mathbf{x}}

\newcommand{\bI}{\mathbf{I}}

\newcommand{\bT}{\mathbf{T}}

\newcommand{\bR}{\mathbf{R}}

\newlength\paramargin
\newlength\figmargin

\newlength\secmargin
\newlength\figcapmargin
\newlength\tabcapmargin

\usepackage{multirow}
\usepackage{float}

\usepackage{tablefootnote}

%% file: sec/0_abstract.tex
\begin{abstract}

We propose \papername, a novel large 3D reconstruction model that leverages a Test-Time Training (TTT) layer to enable long-context, autoregressive 3D reconstruction with linear computational complexity, further scaling the model’s capability. Our framework efficiently compresses multiple image observations into the fast weights of the TTT layer, forming an implicit 3D representation in the latent space that can be decoded into various explicit formats, such as Gaussian Splats (GS) for downstream applications. The online learning variant of our model supports progressive 3D reconstruction and refinement from streaming observations. We demonstrate that pretraining on novel view synthesis tasks effectively transfers to explicit 3D modeling, resulting in improved reconstruction quality and faster convergence. Extensive experiments show that our method achieves superior performance in feedforward 3D Gaussian reconstruction compared to state-of-the-art approaches on both objects and scenes. 

\end{abstract}

%% file: sec/1-intro.tex
\vspace{-15pt}
\section{Introduction}
\label{sec:intro}

Reconstructing explicit 3D representations for photo-realistic rendering from streaming visual input is a central goal of 3D reconstruction. This process is similar to how humans perceive the physical world: we observe a continuous visual stream, build an abstract internal representation of the world, and decode this abstraction into explicit 3D only when needed for fine-grained tasks or to recall detailed 3D structure. In light of this human-like process, 
we aim to enable long-context, autoregressive reconstruction of explicit 3D from streaming visual input.

However, existing 3D reconstruction methods are not designed for long-context scenarios with a memory mechanism.
Traditional approaches to generating 3D representations and synthesizing novel views, including Neural Radiance Fields (NeRF)~\cite{mildenhall2021nerf, yu2021pixelnerf} and 3D Gaussian Splatting (3DGS) ~\cite{kerbl20233d} have achieved substantial progress for high-quality rendering, but they either require slow scene-specific optimization or rely on feedforward reconstruction models with limited input-view scalability.

For example, Large Reconstruction Models (LRMs) have been proposed to rapidly reconstruct various 3D representations such as NeRFs~\cite{hong2023lrm}, meshes~\cite{wei2024meshlrm}, and 3DGS~\cite{zhang2024gs} from input images. However, these models are typically restricted to only a few input views (e.g., four), which limits their ability to reconstruct large-scale scenes. While Long-LRM~\cite{ziwen2025long} extends the number of input views to 32, its use of bidirectional attention layers hinders further scalability and prevents efficient processing of inputs with longer and streamed context, limiting its applicability in real-world scenarios. 
On the other hand, recent research~\cite{sajjadi2022scene, flynn2024quark, jin2024lvsm, zhang2025test} on implicit latent-space 3D representations has demonstrated superior novel view synthesis quality using purely neural networks. However, despite being feedforward models for reconstruction, their rendering speed is significantly slower than that of explicit representations such as 3DGS due to repetitive network inference, and they lack controllability and interpretability, making them less suitable for many downstream applications. 

In this paper, we propose \papername, a novel reconstruction model that leverages neural architectures and the knowledge distilled from pretrained implicit latent-space 3D models, decoding them into explicit 3D representations. This design ensures high-quality novel view synthesis with long-context and autoregressive modeling, while maintaining real-time rendering capability via explicit 3D outputs. 

Our model builds upon Test-Time Training (TTT)~\cite{sun2024learning, zhang2025test} and introduces an architecture composed of LaCT~\cite{zhang2025test} blocks that has only linear computational complexity. We interpret the fast weights of TTT models, which are updated according to inputs during inference, as implicit latent-space 3D representations that can be decoded into various explicit formats such as 3DGS or NeRFs. We demonstrate that, with minimal architectural modification, our framework effectively leverages the pretrained knowledge of large novel view synthesis models~\cite{jin2024lvsm, zhang2025test} for explicit 3D reconstruction. Specifically, our model is trained to \textit{query} the fast weights for different 3D representations, such as a set of virtual view planes for 3DGS, or a triplane feature grid for NeRF-based reconstruction. This design unlocks greater flexibility in the final 3D representation. 
Also, by redesigning the fast-weight update and query mechanism, \papername enables autoregressive 3D reconstruction and refinement with streaming inputs. We further introduce sequence parallelism to enhance scalability.

We validate our model on both object- and scene-level datasets. 
Across both datasets, our model achieves superior reconstruction quality compared to baseline methods, while also being highly efficient. 
We also show that our model supports autoregressive reconstruction, enabling practical real-world applications.
The contributions of our paper can be summarized as following:
\begin{itemize}
\item We propose \papername, the first large reconstruction model that leverages TTT for both feedforward long-context and autoregressive 3D modeling with linear complexity.
\item We design a scalable, unified 3D modeling framework that interprets TTT fast weights into observable and controllable explicit 3D representations.
\item We achieve state-of-the-art results on both object- and scene-level datasets, delivering superior quality and efficiency in 3D reconstruction and novel view synthesis.
\end{itemize}

%% file: sec/2-related-work.tex
\section{Related Work}
\label{sec:related}

\topic{Multi-view 3D Reconstruction}
3D reconstruction from images has been extensively studied in computer vision. Traditional methods such as structure-from-motion~\cite{schonberger2016structure} or multi-view stereo (MVS)~\cite{goesele2006multi} focus on recovering 3D geometry. Deep learning has enabled feed-forward 3D reconstruction~\cite{yao2018mvsnet, yao2019recurrent}, which builds cost volumes using plane sweep for per-view depth estimation. Recently, learning-based MVS approaches~\cite{wang2024dust3r, yang2025fast3r, leroy2024grounding, wang2025vggt, chen2025ttt3r, lan2025stream3r} directly estimate point clouds from input images and have been applied to camera pose estimation. Test3R~\cite{yuan2025test3r} optimizes the network at test time in a self-supervised manner to improve 3D reconstruction. Concurrent work TTT3R~\cite{chen2025ttt3r} defines a gradient to update states for point cloud reconstruction. However, none of these methods can produce photo-realistic novel view synthesis.
 
Neural representations then have emerged as a promising way for both geometry reconstruction and NVS. NeRF~\cite{mildenhall2021nerf} represents the scene as a continuous field and leverages a coordinate-based MLP to predict per-point color and density, enabling differential volumetric rendering with rendering-based supervision. Original NeRF takes hours to optimize a single scene and following works improved its training and rendering efficiency using advanced representations, including voxels~\cite{liu2020neural, sun2022direct}, points~\cite{xu2022point}, hash grids~\cite{muller2022instant}, and triplanes~\cite{gao2023strivec, chan2022efficient, chen2022tensorf}. Recently, 3D Gaussian Splatting~\cite{kerbl20233d, huang20242d} has become the state-of-the-art neural scene representation. It uses volume rendering and rendering loss for optimization similar to NeRF but represents the scene with simple Gaussian primitives which enables real-time rendering and large-scale scene reconstruction~\cite{liu2024citygaussian, kerbl2024hierarchical}.
However, 3DGS still requires optimizing 3D Gaussians from scratch, taking several minutes per scene, whereas our model performs 3D reconstruction within seconds in a feed-forward way.

\topic{Learning-based Feedforward 3D Reconstruction} The development on learning-based methods enables 3D reconstruction and novel view synthesis by training neural networks on large-scale datasets to directly infer 3D structures without per-scene optimization. Early work utilizes Convolutional Neural Networks (CNN) to predict multi-plane images~\cite{flynn2019deepview, mildenhall2019local}, points~\cite{aliev2020neural, yifan2019differentiable} or voxels~\cite{sitzmann2019deepvoxels}.
Large Reconstruction Models (LRM)~\cite{hong2023lrm} propose a transformer-based architecture without 3D inductive bias for 3D object reconstruction from multi-view images, with triplane as the 3D representation.
GS-LRM~\cite{zhang2024gs} further extends LRM to predict pixel-aligned 3DGS, but the model can only take very few images as input due to the quadratic complexity of attention layers. 
Similarly, the subsequence approach~\cite{tang2024lgm, charatan2024pixelsplat, xu2025depthsplat, chen2024mvsplat} also apply a feedforward framework with different neural architectures and 3D inductive bias for Gaussian prediction.
Mamba-based models~\cite{yi2024mvgamba, shen2025gamba} has attempted to reduce the complexity of attention layers, but are still limited to very few input views.
Long-LRM~\cite{ziwen2025long} represents the state of the art in long-sequence Gaussian reconstruction, but it remains limited to 32 input views and relies on additional attention layers. By leveraging TTT, our model achieves longer-context and autoregressive reconstruction with improved NVS quality.

\topic{Linear Attention and State Space Models} To circumvent the quadratic complexity of attention~\cite{vaswani2017attention}, recent research has explored efficient alternatives that retain contextual expressivity while reducing computational cost. Linear attention models~\cite{katharopoulos2020transformers, shen2021efficient, schlag2021linear} approximate the softmax kernel with linearized feature maps to achieve linear complexity, but uniform compression of past key–value pairs often degrades the upper bound of long sequence modeling.

State Space Models (SSMs) introduce a state variable to represent historical information, similar to classical Recurrent Neural Networks (RNNs). Recent works~\cite{sun2023retentive, gu2021efficiently, dao2024transformers, liu2024vmamba} incorporate attenuation factors into the state updates, allowing the model to retain more recent information while gradually forgetting the distant past.
Among them, Mamba~\cite{gu2021efficiently, dao2024transformers, liu2024vmamba} proposes ``date-dependent decay" to model sequences as continuous-time dynamical systems governed by state transition, but it still cannot compete with transformers in long-context reasoning~\cite{waleffe2024empirical}. Jamba~\cite{lenz2025jamba} implements a hybrid mamba attention model to improve the performance.
Test Time Training (TTT), on the other hand, ~\cite{sun2024learning, zhang2025test, behrouz2024titans} transforms the problem into an online learning problem and applies modern optimizers to learn the states. DeltaNet~\cite{schlag2021linear, yang2024parallelizing} and MesaNet~\cite{von2025mesanet} share the same idea but use different update rules when updating.
Inspired by its success, we introduce Test-Time Training into 3D reconstruction tasks for high-quality long-context novel view synthesis, but with only linear complexity.

%% file: sec/3-method.tex
\begin{figure*}[htbp]
    \centering
    \includegraphics[width=0.95\linewidth]{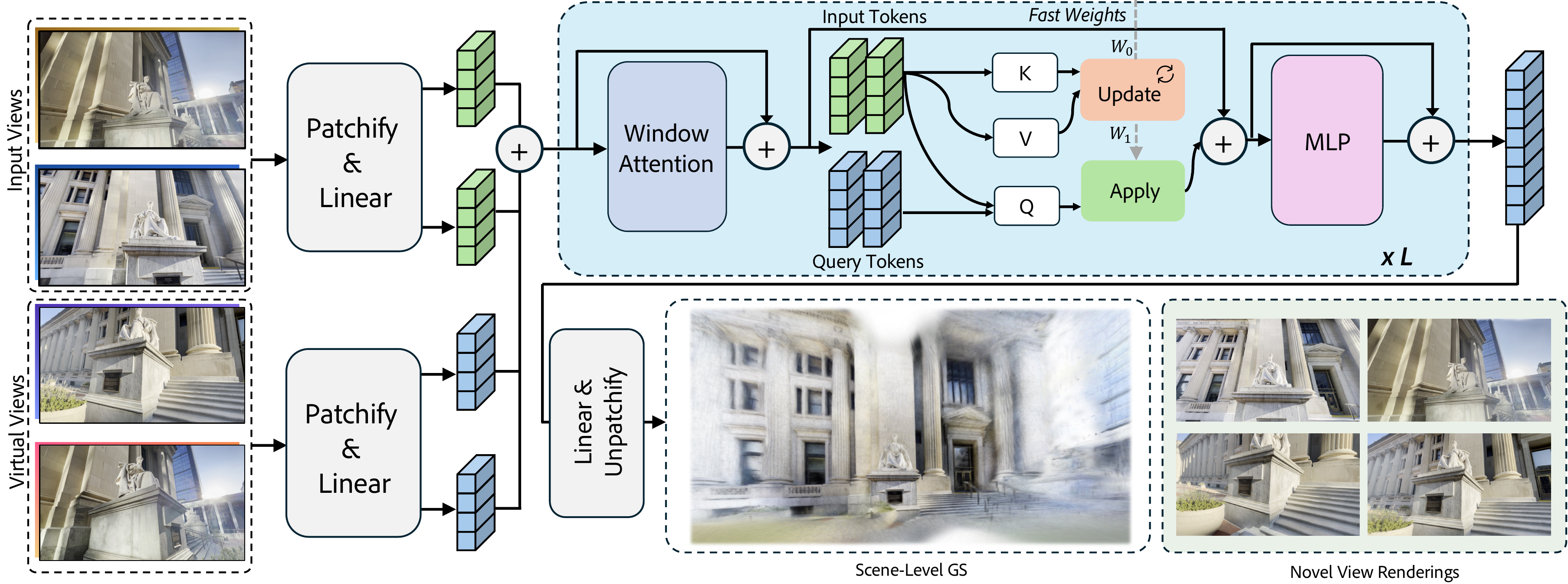}
    \vspace{-10pt}
    \caption{Given a set of posed input images, \papername encodes them into tokens (green boxes) after patchifying. The input tokens are fed into the LaCT block (shown in the blue frame) where fast weights are updated accordingly. Another set of virtual tokens (blue boxes) are used to query the updated fast weights, and decoded into 3D representations like 3DGS for high-quality novel view synthesis.}
    \vspace{-18pt}
    \label{fig:pipeline}
\end{figure*}

\vspace{-8pt}
\section{Method}
\subsection{Preliminary: TTT and LaCT Layer}
We first briefly introduce the fundamentals of TTT and Large Chunk Test-Time Training (LaCT) layer, which form the core building blocks of our model. In sequence modeling, the input is typically represented as a sequence of tokens of length $L$, denoted by $[\bx_1, \bx_2, ..., \bx_L]$, where each token has dimension $d$: $\bx_i \in \mathbb{R}^{d}$. In standard attention, each input token will be projected into query, key and value vectors, denoted as $q_i$, $k_i$, $v_i$. Each token attends to all others via a dot-product operation, leading to quadratic complexity in sequence length. 

TTT~\cite{sun2024learning} learns a set of \textit{fast weights} $W$ that are updated at inference time according to the input to capture the relationship between input tokens. Specifically, it treats the key-value pairs ($k_i$, $v_i$) of input tokens as training data to update fast weights using mean-square error: $W \leftarrow W - \eta \nabla \mathcal{L_{\text{MSE}}}(f_{W}(k), v)$, which can be further applied to queries to obtain the final input $o = f_{W}(q)$. In this way, the fast weights effectively encode the key–value (KV) cache of the input sequence into a fixed-size neural memory. 

Originally, the TTT model~\cite{sun2024learning} updates the fast weights using only a small minibatch (\eg 16 tokens), which results in very low GPU FLOP utilization and difficulty in handling long sequences. 
Large Chunk Test-Time Training (LaCT)~\cite{zhang2025test} instead updates fast weights with large chunk size (up to 1M tokens). Its chunk-wise update computes the gradient of the summed loss over all keys and values within the chunk. 
More details can be found in~\cite{zhang2025test}.

\subsection{Model Architecture}\label{sec:model_arch}
We illustrate our model architecture in \Cref{fig:pipeline}, using 3DGS reconstruction as an example, though the same framework can be applied to other 3D representations as well. Given a set of posed images, denoted as $\{\bI_i \in \mathbb{R}^{H\times W \times 3} | i = 1, 2, .., N\}$, we concatenate them channel-wise with their ray embeddings $\{\bR_i \in \mathbb{R}^{H\times W \times 9} | i = 1, 2, .., N\}$ as the positional embedding. After dividing each image into non-overlapping patches of size $p \times p$, we tokenize these image patches using a lightweight linear layer into a sequence of tokens $\bT$:
\vspace{-2mm}\begin{equation*}
\{\bT_{i,j}\}_{i=1}^N{}_{j=1}^{HW/p^2} = \text{Tokenize}\big( \text{Patchify}([\{\bI_i\}_{i=1}^{N}, \{\bR\}_{i=1}^{N}]) \big),
\label{eq:image_tokenizer}
\end{equation*}
These visual tokens then iteratively update the fast weights $W$ of a set of LaCT blocks using Muon~\cite{jordan6muon} optimizer:
\begin{align} 
    \bT_i &= \bT_i + \text{WinAttn}(\bT_i), \\
    W &= \text{Update}(\{\bT_i\}_{i=1}^{N}), \label{eq:update} \\
   \bT_i &= \text{Apply}(W, \bT_i)\label{eq:apply1}
\end{align}
Each LaCT layer includes a window attention module that captures local relationships within each view. We omit the feedforward layers in the block in the equation for simplicity. The update and apply operations are in linear complexity with respect to the sequence length. 

To retrieve information from the fast weights, we introduce a set of virtual tokens that serve as queries to our model. In 3DGS reconstruction, these virtual tokens are virtual views $\{\bI_i^{\text{v}} \in \mathbb{R}^{H\times W \times 3} | i = 1, 2, .., M\}$ for GS prediction, which will also be patchified and tokenized to $\{\bT_{i,j}^{\text{v}}\}_{i=1}^N{}_{j=1}^{HW/p^2}$. In other 3D representations, such as triplane NeRFs, these virtual tokens are learnable triplane features. The virtual tokens are only used in the apply operation without updating the fast weights:
\begin{align}\label{eq:apply2}
   \bT_i^{\text{v}} &= \text{Apply}(W, \bT_i^{\text{v}}) 
\end{align}

Given the updated query tokens $\bT_i^{\text{v}}$, a linear token decoder transforms them into explicit 3D representations, such as per-patch Gaussian parameters in 3DGS reconstruction. The RGB color, scale, rotation, and opacity of each Gaussian are predicted directly. For Gaussian positions, we first decode the depth of each pixel and use a range function (object-centric for object data and linear for scene data) to convert it to real depth. After that, we convert depth to a Gaussian position with known ray locations and directions.

\vspace{-4pt}
\subsection{Autoregressive Reconstruction} \label{sec:ar_recon}
\vspace{-10pt}
\begin{algorithm}
\caption{Autoregressive 3DGS Reconstruction}
\label{alg:streaming}
\begin{algorithmic}[1]
\REQUIRE Reconstructor $\mathcal{F}$ with initial fast weights $W_0$; input/query view batches 
$\{(\mathcal{I}_{(b)}, \mathcal{I}^{v}_{(b)})\}_{b=1}^{B}$
\ENSURE Reconstructed GS $G$
\STATE $W \leftarrow W_0$
\FOR{$b = 1$ to $B$}
    \STATE $\_, W \leftarrow \mathcal{F}(W, \mathcal{I}_{(b)})$
    \STATE $G_{(b)}, \_ \leftarrow \mathcal{F}(W, \mathcal{I}^{v}_{(b)})$
\ENDFOR
\RETURN $G_{(B)}$
\end{algorithmic}
\end{algorithm}

An important feature of our architecture is its support for autoregressive modeling with streamed input images. To enable this, we modify the update and apply steps 
to incorporate causal dependencies among tokens.
Unlike the standard setting where all input views are jointly processed to update the fast weights before decoding, the streaming variant performs incremental updates in a causal manner.
As illustrated in \Cref{alg:streaming}, providing our model $\mathcal{F}$, for each incoming mini-batch of views $\mathcal{I}_{(b)}$ (\eg, four images at a time), the model updates the fast weights and immediately predicts the corresponding 3D Gaussian parameters for the new query views $\mathcal{I}^{v}_{(b)}$, returning the current reconstructed Gaussian splat results $G_{(b)}$.
This design effectively transforms the model into an RNN-like inference process, where the internal state (fast weights) evolves as new observations arrive, enabling online 3D Gaussian reconstruction.
The fast weight update can also consider historical gradients and fast weights to mitigate drifting (See Supplemental).

\subsection{Distributed Feedforward Reconstruction}
\label{subsec:ddp}
A large number of input views and high-resolution images introduce a substantial number of tokens, leading to a significant increase in both computation and memory cost. A key limitation of prior works lies in their inability to handle long input sequences efficiently, largely due to the lack of parallelism at the sequence level and most methods process all input views within a single device.

\begin{figure}
    \centering
    \includegraphics[width=0.8\linewidth]{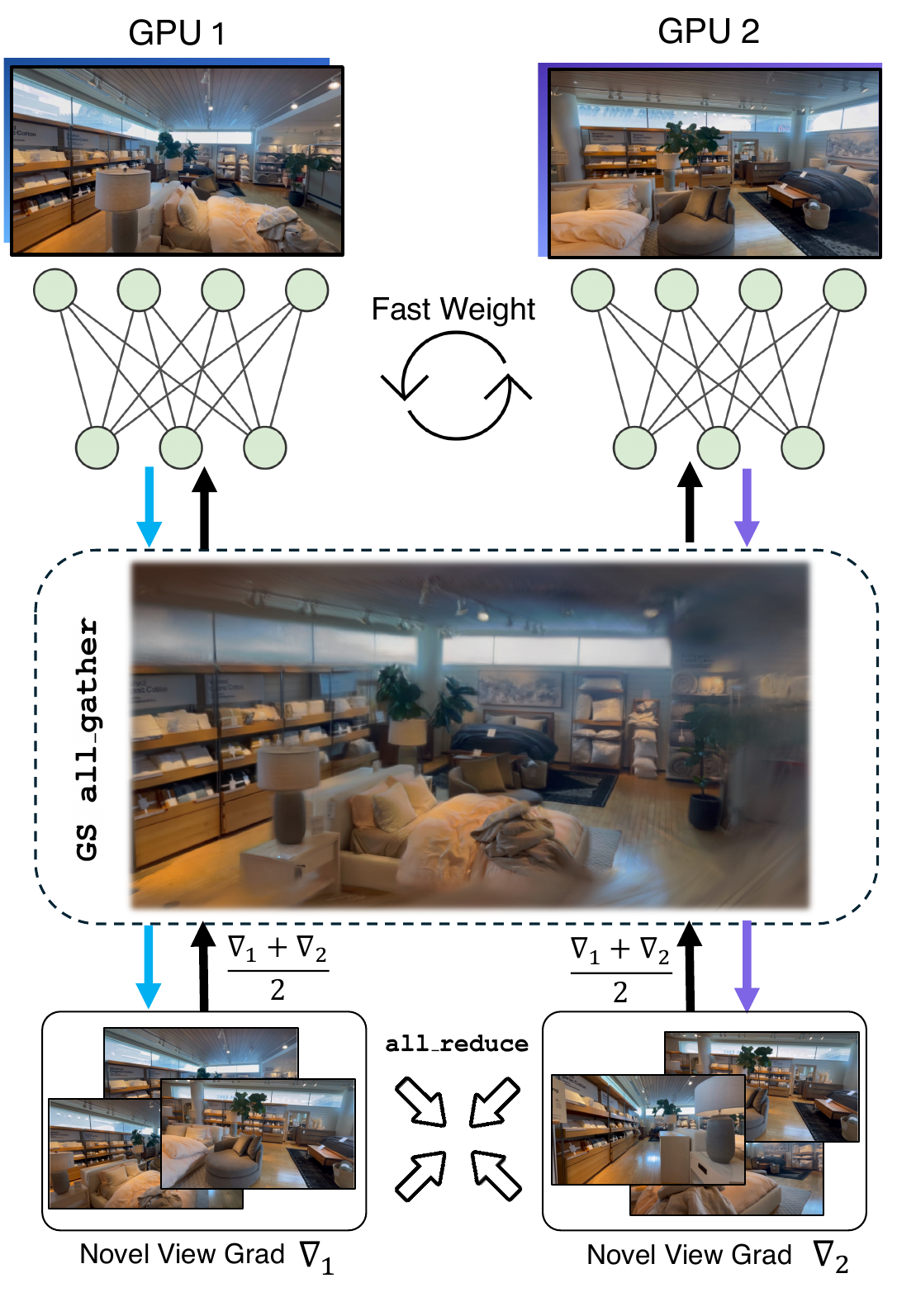}
    \vspace{-8pt}
    \caption{Illustration of distributed feedforward reconstruction training. First, image tokens are sharded across GPUs, and each GPU predicts Gaussians for its assigned virtual views after the fast weights are synchronized. The predicted Gaussians are then gathered to construct the full scene, after which each GPU renders a subset of novel views and computes its respective losses. Gradients are finally all reduced and backpropagated across all devices.}
    \vspace{-12pt}
    \label{fig:sq}
\end{figure}

To address this limitation, we introduce sequence parallelism for training feedforward reconstruction models, exemplified by 3DGS reconstruction as shown in \Cref{fig:sq}. Specifically, we partition the tokenized input views along the sequence dimension and assign each shard to a separate device. During training:
\begin{itemize}
\item Since Gaussians can be predicted independently for each virtual view once the fast weights are synchronized, each GPU predicts pixel-aligned Gaussian primitives for its assigned views (first row).
\item The predicted Gaussians from all devices are gathered to form the complete scene representation (second row).
\item Each GPU subsequently renders its own set of novel views and computes photometric reconstruction losses against the ground truth, and gradients are all reduced to enable sequence-level backpropagation (third row).
\end{itemize}
Thanks to the linearity of our LaCT fast-weight updates, gradients of the fast weights across devices can be easily synchronized through PyTorch Distributed Data Parallel (DDP), ensuring consistent global optimization.
During inference, the distributed reconstruction also allows us to accelerate the reconstruction with more GPUs.

\subsection{Training Objective}
Our training does not require explicit 3D supervision. We render the reconstructed GS on the target views for supervised training, and minimize the rendering loss that is a combination of Mean Squared Error (MSE) and perceptual loss based on VGG-19 features~\cite{simonyan2014very}:
\begin{equation}
    \mathcal{L}_{\text{RGB}} = \text{MSE}(\bI_{\text{pred}}, \bI_{\text{gt}}) + \lambda\:\text{Perceptual}(\bI_{\text{pred}}, \bI_{\text{gt}})
\end{equation}
For non-autoregressive training, we randomly sample unordered input–target image pairs from the dataset. For the autoregressive model, we instead sample ordered input sequences to better simulate streaming use case.

Apart from rendering loss, for scene-level data, we use depth regularization with the scale-invariant depth loss \cite{ziwen2025long} by aligning the Gaussian position along the depth direction (z axis) with ground truth depth for that Gaussian.
We opt for using the monocular depth estimator~\cite{wang2025moge} for pseudo ground truth since we found that feedforward MVS methods like VGGT~\cite{wang2025vggt} provide less detailed depth prediction, albeit being multi-view consistent.
Similar to Long-LRM~\cite{ziwen2025long}, we also use opacity regularization to reduce the number of Gaussians. Our final loss function can be written as follows:
\vspace{-10pt}

\begin{equation}
    \mathcal{L} = \mathcal{L}_{\text{RGB}} + \lambda_{\text{depth}}\mathcal{L}_{\text{depth}} + \lambda_{\text{opacity}}\mathcal{L}_{\text{opacity}}
\end{equation}

\begin{figure*}[htbp]
    \centering
    \includegraphics[width=\linewidth]{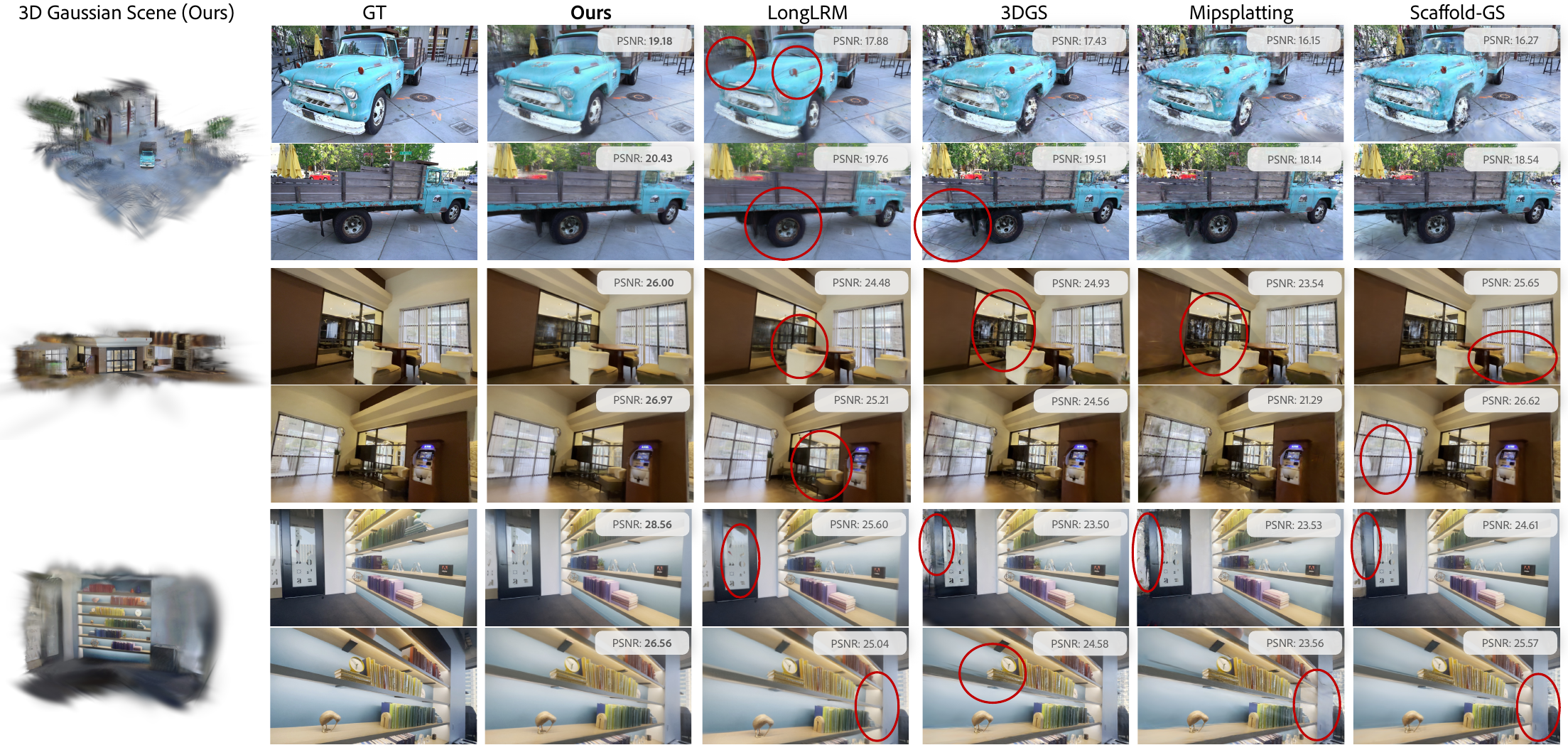}
    \vspace{-20pt}
    \caption{Qualitative comparison between our method and baseline approaches. Our model reconstructs the 3DGS scene with higher fidelity than both optimization-based and feedforward baselines, as also reflected in the PSNR metrics. Please zoom in for a better comparison.}
    \vspace{-10pt}
    \label{fig:results-scene}
\end{figure*}

\begin{figure*}[htbp]
    \centering
    \includegraphics[width=\linewidth]{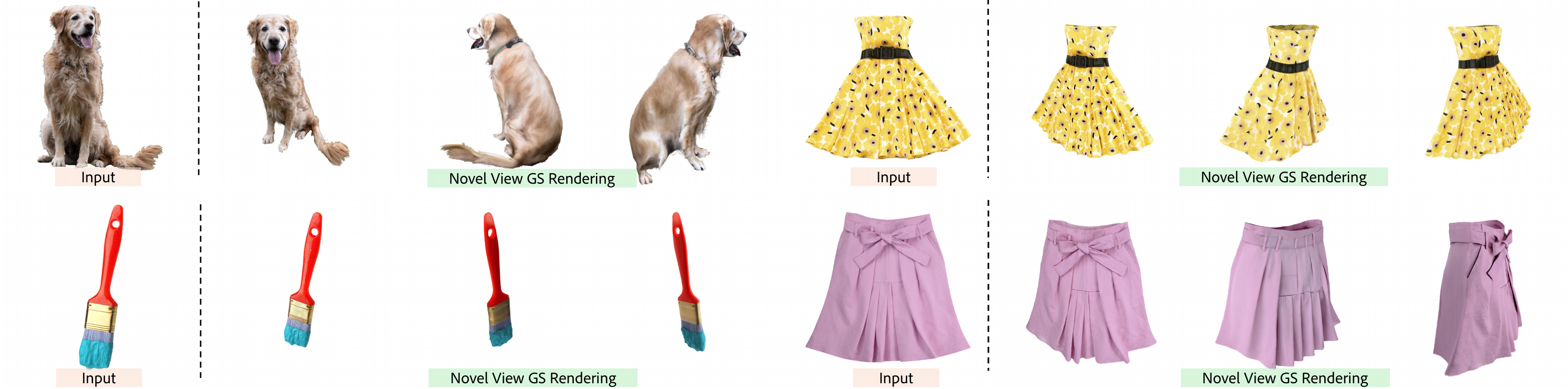}
    \vspace{-20pt}
    \caption{We demonstrate that our high-resolution $1024 \times 1024$ 3DGS \papername can be effectively used for image-to-3D generation when combined with a multi-view generator. Our model enables the reconstruction of fine-grained, photorealistic details \eg, hair, fur, and text, from the input images. Video results are provided in the supplemental material.}
    \vspace{-15pt}
    \label{fig:object1024}
\end{figure*}

\begin{figure*}[htbp]
    \centering
    \includegraphics[width=\linewidth]{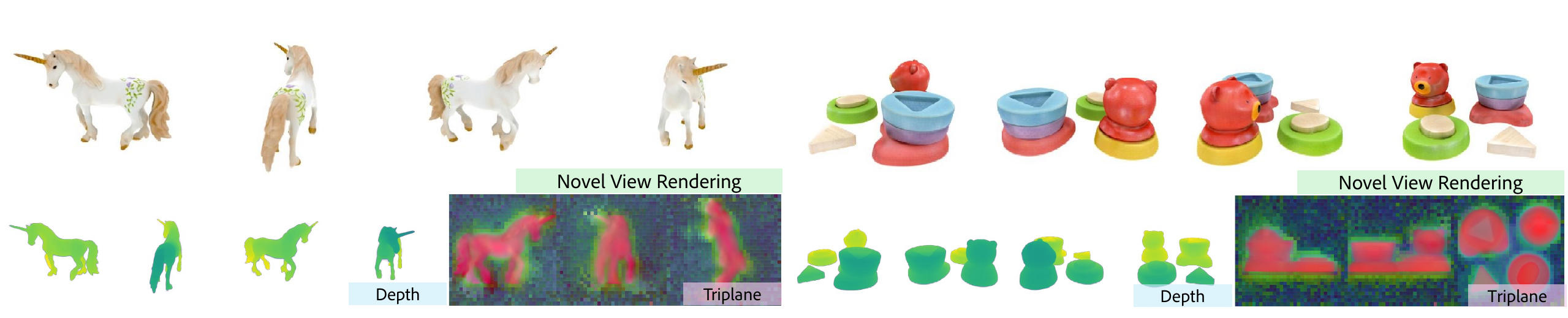}
    \vspace{-20pt}
    \caption{We show that \papername, as a general framework, can also interpret the latent 3D memory into formats besides 3DGS. In this experiment, we use a set of triplane tokens to query the fast weights and then fine-tune the model for triplane-based NeRF reconstruction. We visualize the resulting triplanes and present the corresponding renderings and depth maps for 4 views at a resolution of $512 \times 512$.}
    \vspace{-15pt}
    \label{fig:triplane}
\end{figure*}

%% file: sec/4-experiments.tex
\vspace{-12pt}
\section{Experiments}
\label{sec:exp}

\subsection{Model and Training}
\topic{Model Details} Our model consists of $24$ LaCT blocks with the hidden dimension of $768$. The window attention layers have 64-dimension for each head with QK-normalization for stability.
For the feedforward layer, we use a two-layer MLP with $4$ intermediate expansion ratios for the intermediate dimension. 
We use a patch size of $8 \times 8$ for the image tokenizer. Our architecture shares the same parameterization as TTT-LVSM \cite{zhang2025test} except for the decoding module, allowing us to effectively leverage its pretrained weights as a strong initialization for our model.
\subsection{Datasets}
\topic{Object-level Dataset} We train our object-level reconstruction model on the Objaverse dataset~\cite{deitke2023objaverse}. Following prior works~\cite{hong2023lrm, zhang2024gs}, each 3D object is centered and normalized to fit within a bounding box of $[-1, 1]$. We render 32 views per object, where cameras are randomly distributed around the object at distances uniformly sampled from $[1.5, 2.8]$. All images are rendered at a resolution of $512 \times 512$ under uniform lighting conditions. In total, we use 730K objects for training. We evaluate our model on 100 objects sampled from the Google Scanned Objects (GSO) dataset.
For evaluation, we select a few views as input and the same random 8 views for testing.

\topic{Scene-level Dataset} We train our model on the challenging DL3DV-10K~\cite{ling2024dl3dv} dataset, which consists of 10,510 high-resolution videos, each containing up to 500 keyframes with camera pose annotation obtained from COLMAP~\cite{pan2024global}. The testing set of DL3DV-140 contains 140 test scenes. We use the same input and target split from that provided by Long-LRM~\cite{ziwen2025long}: the testing views are evenly selected from every 8 views (around 40 images each scene) and input views are selected based on K-means clustering based on camera positions and view directions. We also tested our model on Tanks\&Temples~\cite{knapitsch2017tanks} dataset.

\subsection{Baselines and Metrics}
\topic{Object-level} We compare our method with GS-LRM~\cite{zhang2024gs}, an attention-based method. We train the model under 8 input views setting with the same iterations of our method.

\topic{Scene-level} 
Previous feedforward reconstruction methods like GS-LRM~\cite{zhang2024gs} cannot be directly extended to long sequence due to the high complexity of attention. Long-LRM~\cite{ziwen2025long} is the only available feedforward method that can handle more than 16 input views. We also include three optimization-based methods: 3DGS~\cite{kerbl20233d}, Mip-Splatting~\cite{yu2024mip} and Scaffold-GS~\cite{lu2024scaffold}.

\topic{Metrics}
For all baselines, in addition to visual comparisons, we report three metrics to evaluate novel view synthesis quality: PSNR, SSIM, and LPIPS~\cite{zhang2018perceptual}.

\vspace{-5pt}
\subsection{Results}
\topic{Object-level} We present quantitative comparison results under varying resolutions and numbers of input views in \Cref{tab:object-res}.
Across all settings, our method consistently outperforms the baselines. At lower resolutions and shorter sequences, our inference speed is comparable to full attention models, as it is primarily determined by MLP operations.
Thanks to the linear complexity of our architecture with respect to sequence length, at a resolution of $512 \times 512$, our model runs twice as fast as attention-based models while achieving over a 1 dB PSNR improvement.
Our model also demonstrates strong generalization ability—when trained with 8 input views, it can be directly applied to 16 or 24 views (last two rows in the table). With longer sequences, inference becomes substantially faster, and rendering quality further improves through test-time training.

Moreover, our model scales seamlessly to $1024 \times 1024$ resolution, whereas GS-LRM encounters out-of-memory issues under high-resolution training. Results in \Cref{fig:object1024} show that our model can achieve high-quality 3D reconstruction of humans, animals, and texts from a single image when combined with a multi-view diffusion model.

\begin{table}[htbp]
\centering
\small
\caption{Comparison between our method and GS-LRM~\cite{zhang2024gs} on the GSO dataset under different resolutions and numbers of input views.
Our method consistently outperforms GS-LRM in both inference speed and reconstruction quality, and also shows strong generalization ability. V. denotes the number of virtual views used to query the fast weight, which equals input views unless noted.}
\vspace{-6pt}
\resizebox{\columnwidth}{!}{%
\begin{tabular}{ccccccc}
    \toprule
     Method & Resolution & Views & Time (s) $\downarrow$ & PSNR$\uparrow$ & SSIM$\uparrow$ & LPIPS$\downarrow$ \\
     \cmidrule{1-1}\cmidrule(lr){2-2}\cmidrule(lr){3-3}\cmidrule(lr){4-4}\cmidrule(lr){5-7}
     GS-LRM~\cite{zhang2024gs} & \multirow{2}{*}{256 $\times$ 256} & 8 & 0.1 & 31.55 & 0.964 & 0.028 \\
     Ours & & 8 & \textbf{0.1}  &  \textbf{33.14} & \textbf{0.972} & \textbf{0.024} \\
     \cmidrule{1-1}\cmidrule(lr){2-2}\cmidrule(lr){3-3}\cmidrule(lr){4-4}\cmidrule(lr){5-7}
     GS-LRM~\cite{zhang2024gs} &  \multirow{6}{*}{512 $\times$ 512} & 8 & 0.7 & 32.83 & 0.969 & 0.029 \\
     Ours & & 8 & \textbf{0.3} &  \textbf{34.02} & \textbf{0.974} & \textbf{0.025} \\
     \cmidrule{1-1}\cmidrule(lr){3-3}\cmidrule(lr){4-4}\cmidrule(lr){5-7}
     GS-LRM~\cite{zhang2024gs} & & 16  & 2.5 & 33.55  & 0.976 & 0.023\\
     Ours & & 16 (10 V.) & \textbf{0.8} &  \textbf{34.67}  & \textbf{0.978} & \textbf{0.022} \\
     \cmidrule{1-1}\cmidrule(lr){3-3}\cmidrule(lr){4-4}\cmidrule(lr){5-7}
     GS-LRM~\cite{zhang2024gs} & & 24  &  5.5 & 33.26 & 0.976 & \textbf{0.022} \\
     Ours & & 24 (10 V.) & \textbf{1.1} &  \textbf{34.80} & \textbf{0.979} & \textbf{0.022}\\
    \bottomrule
\vspace{-20pt}
\end{tabular}
}
\label{tab:object-res}
\end{table}

\begin{table}[htbp]
\centering
\small
\setlength{\tabcolsep}{4pt}
\renewcommand{\arraystretch}{1.2}
\caption{Quantitative comparison on both DL3DV-140 and Tanks\&Temples datasets under different numbers of input views. Our method surpasses previous feedforward methods and is comparable with optimization-based methods. \textbf{Note that Long-LRM trains a separate model for each input view, while we are a single model across all input views.} Our model can be \textbf{linearly accelerated with multiple GPUs}, here we report time on 1 A100.}
\vspace{-10pt}
\resizebox{\columnwidth}{!}{%
\begin{tabular}{ccccccccc}
\toprule
\multirow{2}{*}{\textbf{Views}} &
\multirow{2}{*}{\textbf{Method}} &
\multirow{2}{*}{\textbf{Time}$\downarrow$} &
\multicolumn{3}{c}{\textbf{DL3DV-140}} &
\multicolumn{3}{c}{\textbf{Tanks\&Temples}} \\
& & & PSNR$\uparrow$ & SSIM$\uparrow$ & LPIPS$\downarrow$ &
PSNR$\uparrow$ & SSIM$\uparrow$ & LPIPS$\downarrow$ \\
\cmidrule{1-1}\cmidrule(lr){2-9}
\multirow{5}{*}{16} 
& 3D GS$_{30k}$ & 13m & 21.20 & 0.708 & 0.264 & 16.76 & 0.598 & 0.334 \\
& Mip-Splatting$_{30k}$ & 13m & 20.88 & 0.712 & 0.274 & 16.82 & 0.616 & 0.332 \\
& Scaffold-GS$_{30k}$ & 16m & \textbf{22.13} & \textbf{0.738} & \textbf{0.250} & \textbf{17.02} & \textbf{0.634} & \textbf{0.321} \\
\cmidrule(lr){2-9}
& Long-LRM (16v model) & \textbf{0.4s} & 22.66 & 0.740 & 0.292 & 17.51 & 0.555 & 0.408 \\
& Ours (single model) & 3.6s & \textbf{23.60} & \textbf{0.784} & \textbf{0.255} & \textbf{18.15} & \textbf{0.613} & \textbf{0.360} \\
\toprule
\multirow{6}{*}{32} 
& 3D GS$_{30k}$ & 13m & 23.60 & 0.779 & 0.213 & 18.10 & 0.688 & 0.269 \\
& Mip-Splatting$_{30k}$ & 13m & 23.32 & 0.784 & 0.217 & 18.39 & \textbf{0.700} & \textbf{0.262} \\
& Scaffold-GS$_{30k}$ & 16m & \textbf{24.77} & \textbf{0.805} & \textbf{0.205} & \textbf{18.41} & 0.691 & 0.290 \\
\cmidrule(lr){2-9}
& Long-LRM (32v model) & \textbf{1s} & 24.10 & 0.783 & 0.254 & 18.38 & 0.601 & 0.363 \\
& Long-LRM (32v model w/ optim) & 12s & 24.99 & 0.809 & 0.243 & 18.69 & 0.623 & 0.360 \\
& Ours (single model, AR) & 7.5s & 24.31 & 0.803 & 0.237 &  18.96 & 0.653 & 0.322 \\
& Ours (single model) & 7.2s & \textbf{25.07} & \textbf{0.822} & \textbf{0.215} & \textbf{19.22} & \textbf{0.662} & \textbf{0.305} \\
\toprule
\multirow{5}{*}{64} 
& 3D GS$_{30k}$ & 13m & 26.55 & 0.852 & \textbf{0.164} & 20.78 & \textbf{0.778} & \textbf{0.205} \\
& Mip-Splatting$_{30k}$ & 13m & 26.29 & 0.850 & 0.166 & 20.08 & 0.759 & 0.220 \\
& Scaffold-GS$_{30k}$ & 16m & \textbf{27.07} & \textbf{0.857} & 0.173 & \textbf{20.96} & 0.768 & 0.240\\
\cmidrule(lr){2-9}
 & Long-LRM (64v model) & \textbf{3.7s} & 24.63 & 0.799 & 0.243 & 19.11 & 0.627 & 0.346 \\
& Ours (single model, AR) & 15.2s &  24.81 & 0.814 & 0.225 & 19.80 & 0.675 & 0.308 \\
& Ours (single model) & 14.8s & \textbf{25.95} &\textbf{0.844} & \textbf{0.195} & \textbf{20.31} & \textbf{0.700} & \textbf{0.274} \\
\bottomrule
\vspace{-32pt}
\end{tabular}
}
\label{tab:scene}
\end{table}

\topic{Scene-level} We further evaluate our model on scene reconstruction, as shown in \Cref{tab:scene}. 
Compared with optimization-based methods that tend to overfit to input views, our method achieves better results on 16 and 32 input views. With more input views, it remains competitive in reconstruction quality, while being hundreds times faster. Moreover, one single \papername model can be applied to different sequence lengths and effectively generalizes to new datasets like Tanks \& Temples.

Compared to the feedforward baseline Long-LRM~\cite{ziwen2025long}, \papername achieves substantially better performance—approximately 1 dB PSNR improvement—across different numbers of input views. 
On the other hand, we show our model constantly outperforms Long-LRM even when it's combined with additional post-optimization. Furthermore, our method can be linearly accelerated by distributing the input across multiple GPUs, as described in \Cref{subsec:ddp}.

\Cref{fig:results-scene} shows visual comparisons between our method and baselines. \papername achieves better visual quality with fewer artifacts than optimization-based methods, thanks to the learned priors across diverse scenes. Our model also outperforms Long-LRM by reconstructing sharper and more detailed geometry (as shown in the red boxes).

\topic{Autoregressive Reconstruction}
We demonstrate the autoregressive reconstruction capability of our model in the second row of \Cref{fig:teaser}. With only 4 input views, the model already produces reasonable 3D Gaussian reconstructions; as additional views arrive (8 and 32 views), both the rendering quality and scene coverage progressively improve. Additional examples of autoregressive reconstruction are provided on our project page.
\Cref{tab:scene} also shows the quantitatively results of our autoregressive model, which constantly outperforms Long-LRM and remains competitive with, or superior to, optimization-based baselines.

\topic{Decoding into Other 3D Formats}
Beyond using 3DGS as the output representation, our architecture can also decode the latent 3D representation into other formats, such as triplane-based NeRFs. As described in \Cref{sec:model_arch}, replacing the virtual tokens with triplane tokens enables the fast weights to be queried as a triplane representation for NeRF reconstruction. We finetune the model with a rendering loss to enable this capability. We show the NeRF renderings and the corresponding queried triplanes in \Cref{fig:triplane}. This demonstrates that our architecture is flexible and can generalize to different 3D output formats.

\subsection{Ablation Study}
We conduct ablation studies to analyze our design choices in LVSM pretraining and the autoregressive reconstruction strategy.

\begin{figure}[htbp]
    \centering
    \includegraphics[width=0.85\linewidth]{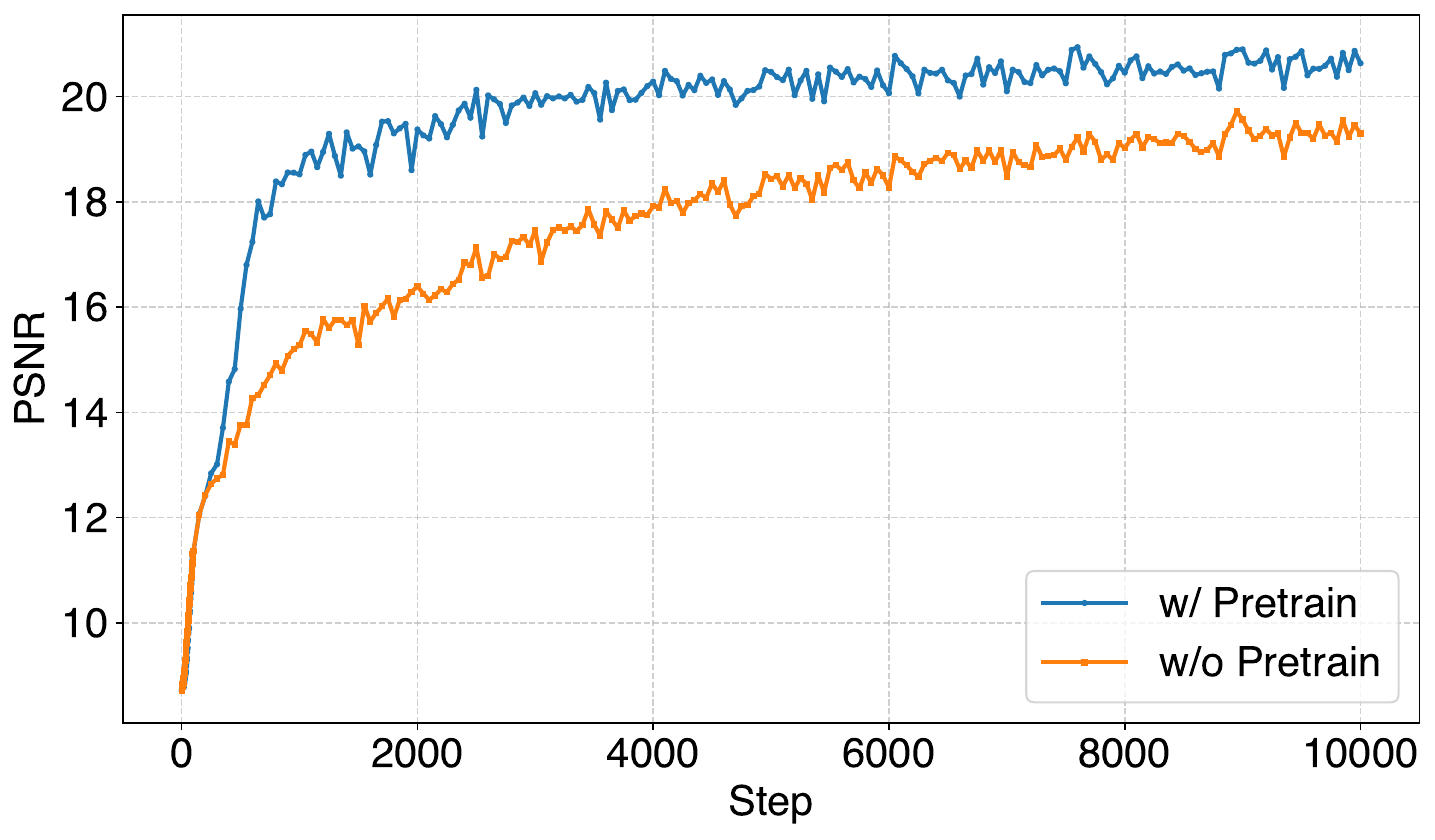}
    \vspace{-12pt}
    \caption{Our 3DGS reconstruction model leverages pretraining with LVSM on novel view synthesis tasks, which significantly accelerates learning and leads to better performance, compared to training from scratch.}
    \label{fig:pretrain}
    \vspace{-12pt}
\end{figure}

\topic{Pretraining from TTT-LVSM} 
We investigate the effectiveness of leveraging pretrained knowledge for both  Gaussian Splatting and triplane training at a resolution of $256 \times 256$. The GS reconstruction has 8 input views with patch size $8 \times 8$, while the triplane version has 4 input views with patch size $16 \times 16$. 
As shown in \Cref{fig:pretrain}, Using GS model as an example, initialization with pretrained checkpoints substantially accelerates convergence, especially in the early training stage, where models quickly reach a high PSNR compared to the one trained from scratch.

Moreover, as reported in \Cref{tab:pretrain}, pretrained initialization not only improves convergence speed but also leads to higher final quality after full training. The gains persist even when trying to adapt the pretrained weights to different 3D representations. The results suggest that pretrained knowledge of novel view synthesis serves as an effective inductive bias for 3D reconstruction, improving both training efficiency and final rendering fidelity.

\vspace{-5pt}
\begin{table}[htbp]
\centering
\small
\caption{Leveraging pretrained knowledge from novel view synthesis tasks improves the final 3D reconstruction quality across different 3D representations.}
\vspace{-8pt}
\begin{tabular}{ccccc}
    \toprule
     3D Rep. & Type & PSNR$\uparrow$ & SSIM$\uparrow$ & LPIPS$\downarrow$ \\
     \cmidrule{1-1}\cmidrule(lr){2-2}\cmidrule(lr){3-5}
     \multirow{2}{*}{GS}
     & w/o Pretrain & 32.77 & 0.026 & 0.969 \\
     &  w Pretrain & \textbf{33.14} & \textbf{0.024} & \textbf{0.972} \\
     \cmidrule{1-1}\cmidrule(lr){2-2}\cmidrule(lr){3-5}
     \multirow{2}{*}{Triplane}
     & w/o Pretrain & 26.40 & 0.903 & 0.093 \\
     & w Pretrain & \textbf{27.87} & \textbf{0.925} & \textbf{0.075} \\
    \bottomrule
\vspace{-25pt}
\end{tabular}
\label{tab:pretrain}
\end{table}

\topic{Autoregressive strategy} In ~\Cref{sec:ar_recon}, we introduce our autoregressive reconstruction strategy. Here we consider a more straightforward way called ``Predict \& Merge": instead of generating a new 3DGS $G_{(b)} \leftarrow \mathcal{F}(W, \mathcal{I}^{v}{(b)})$ for each step, we reuse the previously predicted Gaussians $G_{(b-1)}$ and merge them with the newly predicted subset $G_{(b')} \leftarrow \mathcal{F}(W, \mathcal{I}^{v}_{(b')})$, forming $G_{(b)} = G_{(b-1)} \cup G_{(b')}$.
Here, $\mathcal{I}^{v}_{(b')}$ is a subset of $\mathcal{I}^{v}_{(b)}$ containing only new virtual views not covered in $\mathcal{I}^{v}_{(b-1)}$.
However, we found that though this approach is computationally more efficient, it cannot correct the accumulated errors in $G_{(b-1)}$, leading to worse results than our proposed full reconstruction method, as shown in \Cref{tab:ablation-ar}.


\begin{table}[htbp]
\centering
\small
\caption{Although progressive GS prediction with merging provides more efficient computation, the reconstruction quality is degraded due to accumulated errors (compared on 32 views under 1K iterations finetuning).}
\vspace{-8pt}
\begin{tabular}{cccc}
    \toprule
     & PSNR$\uparrow$ & SSIM$\uparrow$ & LPIPS$\downarrow$ \\
     \cmidrule{1-1}\cmidrule(lr){2-4}
     Predict \& Merge & 21.50 & 0.891 & 0.318 \\
     Ours & \textbf{23.63} & \textbf{0.904} & \textbf{0.259} \\
    \bottomrule
\end{tabular}
\vspace{-15pt}
\label{tab:ablation-ar}
\end{table}

\topic{Optimizer and Losses} We use Muon optimizer for its stabilty and robustness. \Cref{tab:muon} shows that use Muon as opitmizer can bring better results even on low resolution setting. It will bring better results on longer sequence (\eg high resolution and more input views). Also, using depth and opacity as regularization can help reduce opaque Gaussians.
\begin{table}[htbp]
\centering
\vspace{-7pt}
\caption{Ablation on $32$ view $256\times144$ input with the same iterations across settings.}
\vspace{-7pt}
\resizebox{\columnwidth}{!}{%
\begin{tabular}{cccccc}
\toprule
Muon & Opacity+Depth & PSNR$\uparrow$ & SSIM$\uparrow$ & LPIPS$\downarrow$ & Opacity$>0.001$ \\
\midrule
\ding{55} & \ding{55} & 20.44 & 0.649 & 0.295 & 96\% \\
\ding{51} & \ding{55} & 20.68 & 0.661 & 0.290 & 97\% \\
\ding{51} & \ding{51} & \textbf{20.76} & \textbf{0.666} & \textbf{0.285} & \textbf{47\%} \\
\bottomrule
\end{tabular}
}
\vspace{-12pt}
\label{tab:muon}
\end{table}

\vspace{-5pt}
\subsection{Discussions and Limitations}
Our fast-weight memory has a fixed size, which may limit its ability to handle highly complex scenarios with extremely large numbers of input views. More discussions can be found in the supplemental. Also, we observe that, compared with the pretrained LVSM model from which we fine-tune, our quality slightly degraded but we have much faster rendering speed and explicit 3D representations for flexible downstream tasks. This might reflect the inherent trade-off between implicit and explicit representations. Future works might design a better memory mechanism, further improve the quality, and speed up the inference to enable real-time high-quality reconstruction for streaming inputs.




%% file: sec/5-conclusion.tex
\vspace{-5pt}
\section{Conclusion}
\label{sec:conclusion}
In this paper, we present \papername, a large reconstruction model that supports both feedforward long-context and autoregressive 3D modeling. Under the Test-Time Training framework, it produces implicit fast-weight representations and converts them into explicit 3D representations such as Gaussian splats and triplanes for efficient, high-quality novel view synthesis. Experiments on object- and scene-level datasets show that \papername outperforms prior feedforward methods in quality and scalability while approaching the speed of explicit representations. Our framework helps close the gap between neural network rendering and real-time explicit 3D systems. 


%% file: sec/X_suppl.tex
\clearpage
\setcounter{page}{1}
\maketitlesupplementary

\etoctoccontentsline{part}{Appendix}
\localtableofcontents


\section{Further Discussions}
\label{sec:discussion}
\topic{Effect of Scene Complexity on Fast Weights}
The memory of fast-weights has a fixed capacity and is bounded, especially in the autoregressive setting.
Our empirical analysis on DL3DV scene labels indicates that higher scene complexity leads to degraded performance, as observed in outdoor vs. indoor scenes (PSNR: 24.45 vs. 24.96) and high- vs. low-frequency scenes (PSNR: 24.20 vs. 25.97).
The memory capacity is also influenced by sequence length, where earlier inputs may be gradually forgotten as more tokens are processed.

\topic{Selective Update of Fast Weights in AR Setting} Instead of updating the fast weights only according to current inputs, we can further use history states for selective update to mitigate drifiting. Inspired by ~\cite{ma2025fast}, we explore a mechanism to prevent weight drift. Specifically, we approximate the diagonal of the Fisher information using an exponential moving average of squared gradients, as an estimate of parameter importance. Meanwhile, we maintain a sliding anchor via EMA to track the historical trajectory of the fast weights. After each gradient update, we apply elastic regularization based on parameter importance. Specifically, we leverage Fisher information for selective update, where parameters with high Fisher values that are important for the current input, are left parameters with high Fisher values unconstrained, while parameters with low Fisher values are pulled back toward the anchor. This encourages adaptation to the current input and suppresses drift in unimportant parameters. This training-free strategy can further improve our autoregressive model, and we envision it to be more effective by incorporating it into training for future work.

\begin{table}[htbp]
\centering
\small
\caption{Training-free selective update considering history fast weights can further enhance our AR model.}
\vspace{-8pt}
\begin{tabular}{cccc}
    \toprule
     & PSNR$\uparrow$ & SSIM$\uparrow$ & LPIPS$\downarrow$ \\
     \cmidrule{1-1}\cmidrule(lr){2-4}
     w/o selective & 24.81 & 0.814 & 0.225 \\
     w selective & \textbf{24.95} & \textbf{0.818} & \textbf{0.223} \\
    \bottomrule
\end{tabular}
\vspace{-10pt}
\label{tab:selective}
\end{table}

\topic{Scaling to More Input Views} With distributed training, \papername can be further scaled to hundreds of views given enough compute. For example, by finetuning our full model with more iterations on $128$ input views (more than 1M tokens), it can achieve $26.80$ PSNR.

\begin{figure}[htbp]
    \centering
    \includegraphics[width=0.85\linewidth]{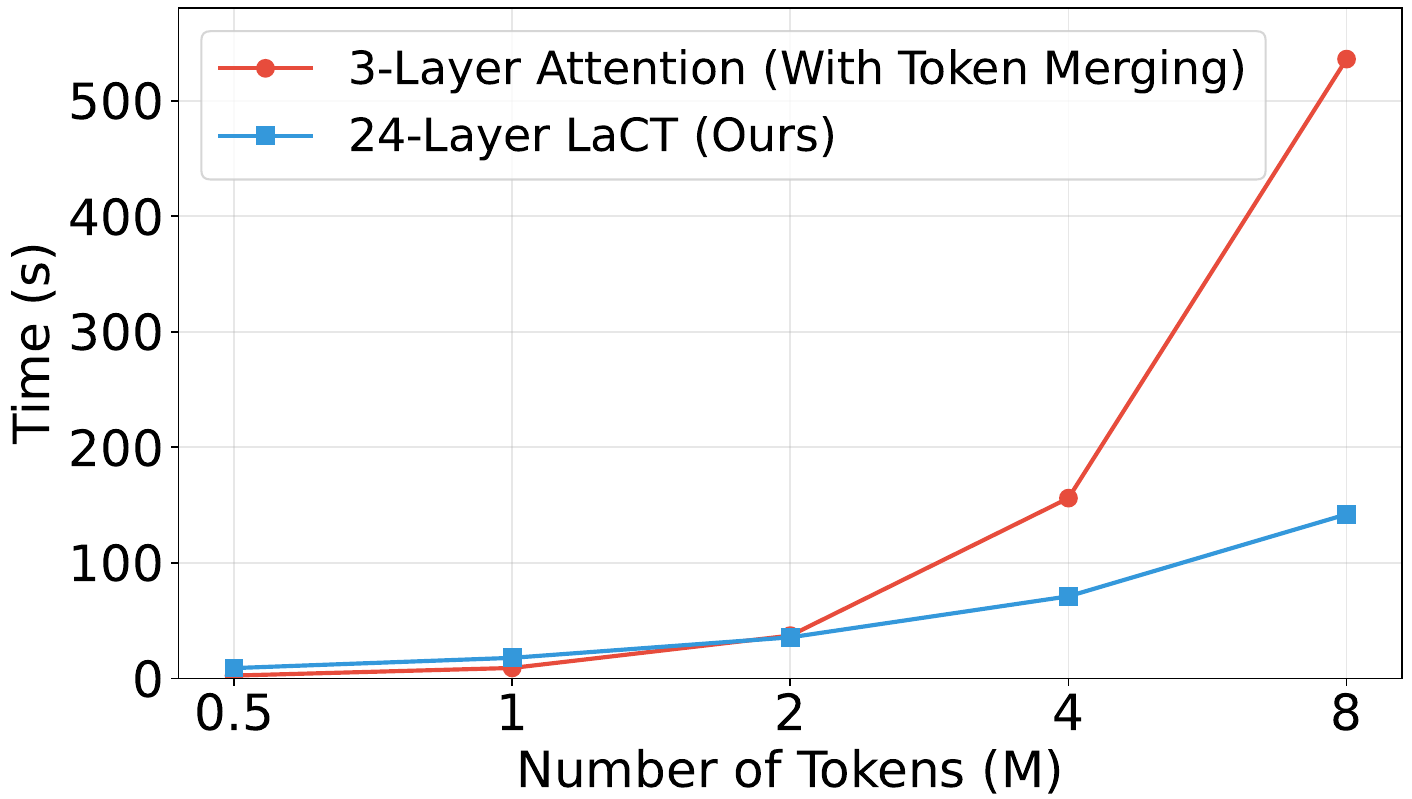}
    \vspace{-8pt}
    \caption{Time comparison of 3 Attention layers vs 24 layers of LaCT blocks under different numbers of tokens.}
    \label{fig:time}
    \vspace{-15pt}
\end{figure}

\topic{Possible Usage of Attention Layers}
We \underline{deliberately avoid} attention blocks in our model since it has quadratic complexity $O(N^2d)$ compared to our linear FLOPS $O(Nd^2)$ of LaCT blocks ($N$ is the number of tokens and $d$ is hidden dimension). Therefore, attention will bottleneck the computation with growing number of tokens and be very slow in our million-level token setting. 
As shown in \Cref{fig:time}, even a 3-layer attention only will be slower than our 24-layer LaCT blocks from 2M tokens (256 views). With more compute, our model can easily scale to longer sequence and remain linear complexity.

\section{Experiment Details}
\label{sec:exp-details}

\subsection{Scene-level Training}

We adopt a curriculum training strategy that progresses from low to high resolution, motivated by two main reasons. First, fast low-res pretraining enables the model to train with a large batch size with faster iteration time. Second, we found that even with pretrained TTT-LVSM~\cite{zhang2025test} checkpoints at high-resolution (\ie $960 \times 540$), the model cannot predict reasonable Gaussians at the beginning iterations, leading to excessive GPU memory usage due to the rendering of a large number of Gaussians.

For scene-level training, We train our model on three stages with $144 \times 256$, $288 \times 512$ and $540 \times 960$ resolution.
For each stage, we resize the images to the target resolution, which all have the same aspect ratio as the original dataset. For all stages, we first determine a continuous range based on the start and end frames from the entire video sequence from the dataset. The range is randomly sampled from $128$ to $512$ for each sample to ensure enough coverage of the scene. Then, we randomly sample $124$ frames from this range, from which both input and target views will be further sampled. They are ensured to have overlap frames for stable training. We train the model across $16$ to $64$ input views. For training, we use the input views as the virtual views and found that provides the best results.

For the first stage, we train the model with a peak learning rate of $3\mathrm{e}{-4}$ with $2$K warmup steps and cosine decay.
We use AdamW optimizer with betas $(0.9, 0.95)$ and weight decay $0.05$.
We train the model using a batch size of $128$ for $80$K steps, which is around $0.3$T tokens.
For the second stage, we finetune the model at the resolution of $288 \times 512$ with a peak learning rate of $5\mathrm{e}{-5}$. We use a batch size of $64$ to train $6$K steps. For the final stage, we enable depth loss and opacity loss, training the model with 32 input views with $5$K steps with peak learning rate $1\mathrm{e}{-5}$ and batch size of $64$. Finally, we train the model with 16 to 64 input views for another $1$K steps. We prune $70\%$ Gaussians with the smallest opacity for 64 views and $60$\% otherwise. 

For autoregressive model training, we finetune our model on the final stage checkpoints for around another $3$K iterations with peak learning rate $1\mathrm{e}{-4}$ and batch size of $64$. We train the model on input views from $8$ to $64$. Our models are trained on 64 Nvidia A100 80GB GPUs.

Besides, we use \texttt{gsplat} Python library for efficient Gaussian training. We enable \texttt{torch.compile} to accelerate computation, achieving roughly a 30\% per-iteration speedup. To further optimize memory and stability, we implement gradient checkpointing~\cite{chen2016training} and mixed-precision training~\cite{micikevicius2017mixed} with the \texttt{BFloat16} format. For Gaussian rendering, we utilize deferred backpropagation~\cite{zhang2022arf} to reduce GPU memory consumption. In addition, iterations with a gradient norm that exceeds 5.0 are skipped to improve training stability.

\subsection{Object-level training}
For the GS-based model, we use 8 views as input and another 8 views as supervision and use a patch size of $16 \times 16$. We firstly sample a set of 15 images (from 32 renderings) as a data point, from which
we randomly select 8 input views and 8 supervision views independently. This sampling strategy encourages more overlap between input views and rendering views than directly sampling from 32 rendering views. We train on the resolution of $256 \times 256$ with a batch size of $512$ for $80$K iterations with a peak learning rate of $4\mathrm{e}{-4}$. We then finetune on $512 \times 512$ with a batch size of $128$ for another $10$K iterations with a peak learning rate of  $5\mathrm{e}{-5}$. We further finetune on $1024 \times 1024$ with a batch size of $64$ for another $4$K iterations with a peak learning rate of  $5\mathrm{e}{-5}$.

For the triplane-based model, we use 4 views as input and another 4 views as supervision and use a patch size of $16 \times 16$. We train on the resolution of $256 \times 256$ with a batch size of $256$ for $60$K iterations and finetune on $512 \times 512$ with a batch size of $64$ for another $20$K iterations.

\section{More results and Comparison}
In \Cref{tab:scene-supp}, we show results where we combine our method with a few additional optimization steps. It demonstrates that the reconstructed model can be further improved with minimal optimization cost, surpassing both purely optimization-based methods and the previous state-of-the-art feed-forward method, Long-LRM, under the same post-optimization setup. 

Notably, the quality of Long-LRM with 3-step post-optimization is still lower than our model \textit{without} post-optimization, even though it requires more time to perform the optimization than our feedforward inference.


\begin{table}[htbp]
\centering
\small
\setlength{\tabcolsep}{4pt}
\renewcommand{\arraystretch}{1.2}
\caption{More quantitative comparison on both DL3DV-140 and Tanks\&Temples datasets under 32/64 input views. Our method surpasses previous feedforward methods and can further surpass optimization-based methods with a few steps post-optimization. \textbf{Note that Long-LRM trains a separate model for each input view, while we are a single model across all input views.} Our model can be \textbf{linearly accelerated with multiple GPUs}, here we report time on a single Nvidia A100 80GB GPU.}
\vspace{-5pt}
\resizebox{\columnwidth}{!}{%
\begin{tabular}{ccccccccc}
\toprule
\multirow{2}{*}{\textbf{Views}} &
\multirow{2}{*}{\textbf{Method}} &
\multirow{2}{*}{\textbf{Time}$\downarrow$} &
\multicolumn{3}{c}{\textbf{DL3DV-140}} &
\multicolumn{3}{c}{\textbf{Tanks\&Temples}} \\
& & & PSNR$\uparrow$ & SSIM$\uparrow$ & LPIPS$\downarrow$ &
PSNR$\uparrow$ & SSIM$\uparrow$ & LPIPS$\downarrow$ \\
\toprule
\multirow{10}{*}{32} 
& 3D GS$_{30k}$ & 13m & 23.60 & 0.779 & 0.213 & 18.10 & 0.688 & 0.269 \\
& Mip-Splatting$_{30k}$ & 13m & 23.32 & 0.784 & 0.217 & 18.39 & \textbf{0.700} & \textbf{0.262} \\
& Scaffold-GS$_{30k}$ & 16m & \textbf{24.77} & \textbf{0.805} & \textbf{0.205} & \textbf{18.41} & 0.691 & 0.290 \\
\cmidrule(lr){2-9}
& Long-LRM (32v model) & \textbf{1s} & 24.10 & 0.783 & 0.254 & 18.38 & 0.601 & 0.363 \\
& Long-LRM (w/ 3-step optim) & 12s & 24.99 & 0.809 & 0.243 & 18.69 & 0.623 & 0.360 \\
& Long-LRM (w/ 10-step optim) & 37s & 25.60 & 0.826 & 0.233 & 18.90 & 0.642 & 0.350 \\
& Ours & 7.2s & 25.07 & 0.822 & 0.215 & 19.22 & 0.662 & 0.305 \\
& Ours (w/ 3-step optim) & 18s & 25.86 & 0.842 & 0.208 & 19.57 & 0.687 & 0.300 \\
& Ours (w/ 10-step optim) & 42s & \textbf{26.37} & \textbf{0.854} & \textbf{0.201} & \textbf{19.78} & \textbf{0.704} & \textbf{0.291} \\
\toprule
\multirow{10}{*}{64} 
& 3D GS$_{30k}$ & 13m & 26.55 & 0.852 & \textbf{0.164} & 20.78 & \textbf{0.778} & \textbf{0.205} \\
& Mip-Splatting$_{30k}$ & 13m & 26.29 & 0.850 & 0.166 & 20.08 & 0.759 & 0.220 \\
& Scaffold-GS$_{30k}$ & 16m & \textbf{27.07} & \textbf{0.857} & 0.173 & \textbf{20.96} & 0.768 & 0.240\\
\cmidrule(lr){2-9}
& Long-LRM (64v model) & \textbf{3.7s} & 24.63 & 0.799 & 0.243 & 19.11 & 0.627 & 0.346 \\
& Long-LRM (w/ 3-step optim) & 38.9s & 25.74 & 0.833 & 0.225 & 19.69 & 0.659 & 0.333 \\
& Long-LRM (w/ 10-step optim) & 114s & 26.72 & 0.852 & 0.212 & 20.03 & 0.681 & 0.320 \\
& Ours & 14.8s & 25.95 & 0.844 & 0.195 & 20.31 & 0.700 & 0.274 \\
& Ours (w/ 3-step optim) & 47s & 26.97 & 0.866 & 0.185 & 20.76 & 0.724 & 0.269\\
& Ours (w/ 10-step optim) & 124s & \textbf{27.65} &\textbf{0.880} & \textbf{0.177} & \textbf{21.07} & \textbf{0.743} & \textbf{0.260} \\
\bottomrule
\vspace{-32pt}
\end{tabular}
}
\label{tab:scene-supp}
\end{table}

%